\documentclass[11pt]{article}
\usepackage[final]{eacl2023}

\usepackage{times}
\usepackage{latexsym}
\usepackage[T1]{fontenc}
\usepackage[utf8]{inputenc}
\usepackage[longtable]{multirow}
\usepackage{longtable}
\usepackage{makecell}
\usepackage{comment}
\usepackage{adjustbox}
\usepackage{enumitem}
\usepackage{amsmath}
\usepackage{booktabs}
\usepackage{colortbl}
\usepackage{diagbox}
\usepackage{pifont}
\usepackage{amssymb}

\usepackage{arydshln}
\usepackage{float}
\usepackage{afterpage}
\usepackage{capt-of}
\usepackage{tabularx}
\usepackage{tabularray}
\usepackage{siunitx}
\usepackage{tcolorbox}
\usepackage[safe]{tipa}
\usepackage{placeins}
\usepackage{flafter}

\usepackage{array}

\newcolumntype{P}[1]{>{\centering\arraybackslash}p{#1}}

\usepackage{makecell}

\usepackage{mathtools}
\usepackage{blkarray, bigstrut} 
\usepackage{microtype}
\usepackage{inconsolata}

\usepackage[normalem]{ulem}

\usepackage{tikz}
\definecolor{northern}{RGB}{207, 239, 102}
\definecolor{central}{RGB}{255, 255, 0}
\definecolor{southern}{RGB}{250, 128, 114}
\definecolor{zaza}{RGB}{255, 215, 0}
\definecolor{gorani}{RGB}{138, 43, 226}
\definecolor{mixed}{RGB}{179, 212, 35}

\usetikzlibrary{calc}
\usetikzlibrary{decorations.pathmorphing}
\usetikzlibrary{shapes,arrows,chains}
\usetikzlibrary{shapes.geometric}
\usetikzlibrary{patterns}
\usetikzlibrary{positioning}
\tikzset{>=latex}
\usepackage{pgfplots}
\usepackage{relsize}

\title{A Factorial Study of Synthetic Data Generation for Low-Resource Machine Translation using Grammar Books}

\author{Varun Ghat Ravikumar$^{1}$ \quad Sina Ahmadi$^{2}$ \quad Lena Jäger$^{2}$ \quad Rico Sennrich$^{2}$ \\
  $^{1}$Department of Informatics, University of Zurich \\
  $^{2}$Department of Computational Linguistics, University of Zurich \\
  \texttt{\{varunghat.ravikumar, sina.ahmadi, lenaann.jaeger, rico.sennrich\}@uzh.ch}}

\begin{document}

\maketitle
\thispagestyle{plain}
\pagestyle{plain}

\begin{abstract}
Most endangered languages lack the parallel data required for machine translation, despite the existence of descriptive grammar books. We introduce a pipeline that uses large language models to extract grammatical rules, example sentences, and lexicons from grammar books and generate synthetic parallel corpora for fine-tuning---rather than feeding grammar content into prompts at inference time, as in prior work. Validated on three typologically diverse low-resource languages---Kalamang (Papuan), Tuatschin (Romance), and Mandan (Siouan)---we show that fine-tuning on synthetic data improves over seed-data baselines in 75\% of configurations for Kalamang and 59\% for Tuatschin, with best-case ChrF++ gains of +8.8, +5.3, and +3.3 respectively. Through a systematic factorial study across 96 configurations varying target part-of-speech, retrieval granularity, and sample volume, we identify which factor combinations drive gains and where they break down. Our results demonstrate that static linguistic documentation can be repurposed for machine translation fine-tuning, offering a practical path towards translation tools for severely under-resourced languages.

\begin{minipage}{\columnwidth}
    \centering
    \raisebox{-0.15cm}{\includegraphics[height=0.5cm]{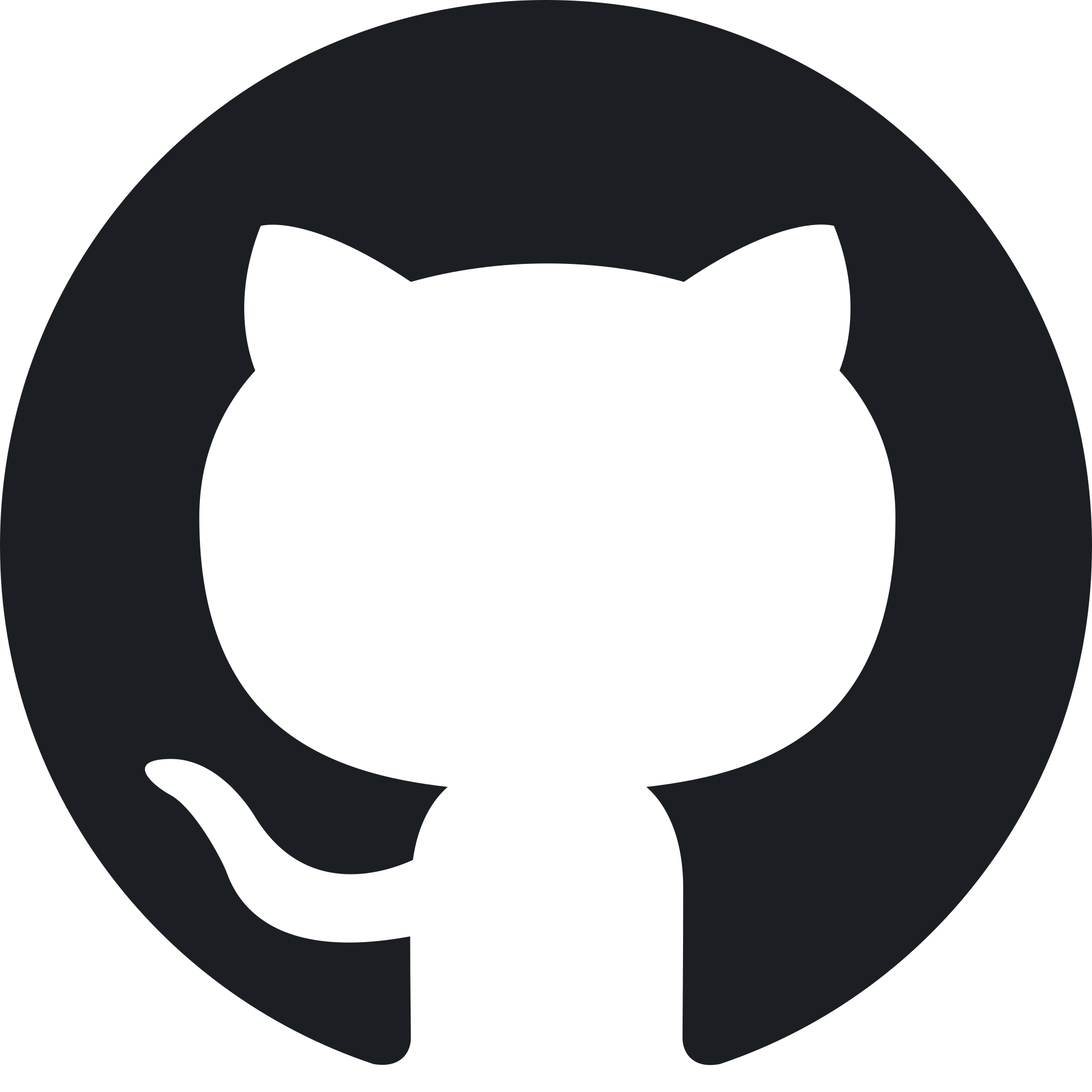}}
    ~\href{https://github.com/varunghat/GrammarMT}{\textbf{varunghat/GrammarMT}}
\end{minipage}

\end{abstract}

\section{Introduction}

Nearly half of the world's 7,000+ languages are endangered, and many may vanish within the next century if preservation efforts do not accelerate \citep{unesco_languages_2022}. While high-resource languages benefit from large-scale monolingual and parallel corpora, most endangered and minority languages lack these basic training data \citep{haddow-etal-2022-survey}. Despite limited digital presence, many endangered languages possess descriptive grammar books compiled through linguistic field work~\cite{DBLP:conf/semweb/NordhoffH11}, yet such data remains largely unexploited in natural language processing (NLP). 

\begin{figure}[t]
    \centering
    \includegraphics[width=1\columnwidth]{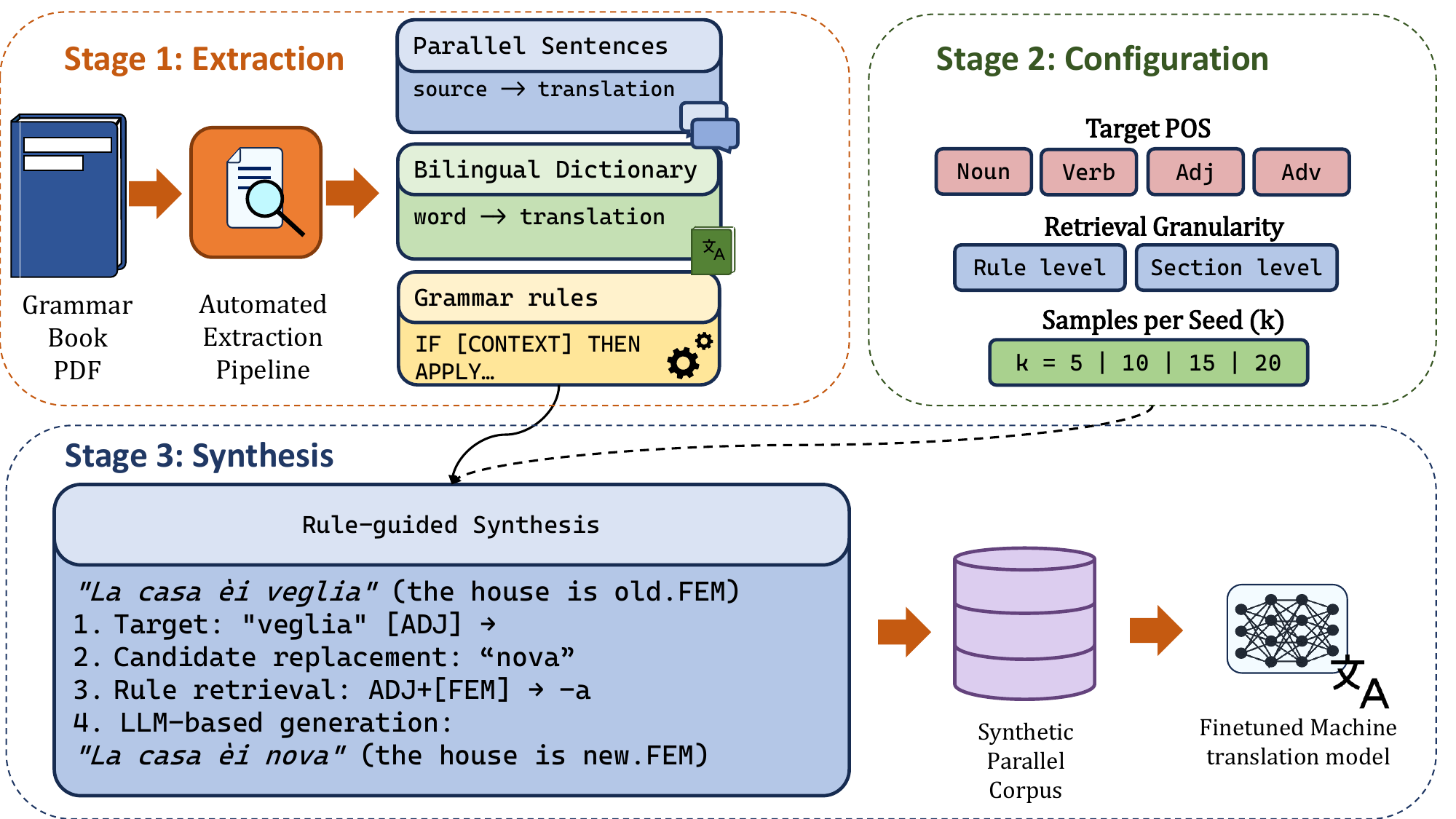}
    \caption{Overview of our methodology. Stage 1 extracts linguistic information (parallel sentences, dictionaries, rules). Stage 2 shows the combinatorial experimental configuration. Stage 3 generates synthetic variants via rule-guided lexical substitution (Tuatschin example shown).}
    \label{fig:intro_pipeline}
\end{figure}

Grammar books document grammatical rules, parallel sentence examples, and lexicons, offering structured descriptions of morphology, syntax, and vocabulary \citep{Leech_2015}. \citet{tanzer2024benchmarklearningtranslatenew} show that prompting LLMs with grammar book content improves translation over zero-shot baselines. However, \citet{aycock2025llmsreallylearntranslate} observe that LLMs primarily benefit from embedded parallel sentences rather than raw grammatical descriptions. In contrast, \citet{zhang2025readstepstranslatingextremely} show that when grammatical rules are manually codified into structured representations, they do improve both translation quality and rule retrieval. Despite the growing interest in leveraging grammar books and reasoning capabilities of large language models (LLMs) for low-resourced NLP~\citep[\textit{inter alia}]{hus2024back,spencer2025can}, the impact of the distinction between narrative descriptions and codified rules has not been thoroughly examined. This motivates our central question: \emph{whether the structured content of grammar books can be operationalized computationally, and use explicit grammatical instructions to generate novel sentences.}

We introduce a semi-automated framework for synthetic data generation from grammar books, requiring only minimal supervision in the form of $\sim$200 annotated lines per language to train the sentence classifier. Automating this process is necessary because manual rule codification requires expert linguistic effort that does not scale to the thousands of languages with existing grammars. Our pipeline, illustrated in Figure~\ref{fig:intro_pipeline}, extracts parallel sentences, word-lists, and grammatical rules from unstructured PDF text. These components are then used to generate synthetic parallel corpora by replacing words in extracted sentence templates while applying retrieved grammatical rules to maintain morphological and syntactic correctness. We evaluate the generated corpora by fine-tuning Gemini-2.5-flash on the synthetic data and measuring translation performance on held-out test sets from three typologically diverse and severely low-resource languages: Kalamang (agglutinative, Papuan), Tuatschin (analytic, Romance), and Mandan (polysynthetic, Siouan). We selected Gemini-2.5-flash for its fine-tuning API availability and low inference cost. Through ablation studies, we analyze the impact of retrieval granularity (rule-level vs.\ section-level), target part-of-speech, and generation volume on translation quality.
\section{Related Work}

Machine translation (MT) performance relies on large-scale parallel corpora. For low-resource languages, massively multilingual models such as NLLB \citep{nllb-team-2022-language} apply cross-lingual transfer from high-resource languages, but performance degrades significantly for languages absent from pre-training data~\citep{ahmadi2025parme}. Back-translation \citep{sennrich-etal-2016-improving} generates synthetic source sentences from monolingual target text but requires an initial translation model of sufficient quality, which is often unavailable for extremely low-resource languages. \citet{frontull-moser-2024-rule} find that combining neural models with explicit rule-based constraints helps preserve structural integrity when parallel data is scarce, motivating the use of grammatical rules in our pipeline.

A separate line of work has explored using descriptive grammar books to support translation for languages without web data. \citet{tanzer2024benchmarklearningtranslatenew} show that LLMs can perform translation when prompted with entire grammar books, though \citet{aycock2025llmsreallylearntranslate} find that models primarily learn from embedded parallel examples rather than descriptions. \citet{hus-anastasopoulos-2024-back} corroborate this finding across 16 languages, observing that grammar book prompting yields only modest gains at high computational cost due to token-heavy context injection. Other work has pursued structured retrieval to reduce context injection costs~\citep{guo-etal-2024-teaching,zebaze-etal-2025-compositional}, while \citet{zhang2024teachinglargelanguagemodels} chain morphological analyzers and dictionaries, requiring pre-existing tools rarely available for low-resource languages. Most recently, \citet{zhang2025readstepstranslatingextremely} convert grammatical rules into executable code to guide translation, achieving strong results but relying on manual expert annotation.

Complementary to direct translation, synthetic data generation provides an alternative augmentation strategy. Prior work has explored lexical substitution scored by language models~\citep{fadaee-etal-2017-data}, learning joint source-target probabilities~\citep{ding-etal-2020-daga}, and combining morphological features with dictionary entries~\citep{alam2024morphologically}. \citet{anikina-etal-2025-rigorous} find that combining target-language demonstrations with LLM-based revision narrows the gap with gold-standard data to within 5\%. At the grammar-guided end, \citet{lucas-etal-2024-grammar} manually construct formal grammars to generate parallel corpora for Guarani, demonstrating the value of linguistic constraints but requiring expert effort. Conversely, \citet{degibert2025scalinglowresourcemtsynthetic} generate massive synthetic corpora without grammar constraints, achieving scale but lacking morphological precision.

Current approaches face a bottleneck: they either inject entire grammar books into context windows at high computational cost with limited utilization of grammatical content \citep{aycock2025llmsreallylearntranslate}, or rely on manual expert annotation of grammatical rules \citep{zhang2025readstepstranslatingextremely}. Furthermore, retrieval-augmented approaches \citep{guo-etal-2024-teaching,zebaze-etal-2025-compositional} operate during inference, which does not address training data scarcity for fine-tuned models. Existing augmentation methods that do generate training data rely on monolingual LM probabilities \citep{fadaee-etal-2017-data} or unconstrained LLM generation \citep{degibert2025scalinglowresourcemtsynthetic}. No existing method automatically parses grammar books with minimal human intervention, specifically to create synthetic training data for machine translation models. Our work fills this gap.
\section{Methodology}
\subsection{Linguistic Information Extraction}

Descriptive grammar books contain parallel sentence examples in heterogeneous formats. A typical example consists visually of one or more source-gloss line pairs where each source line is immediately followed by its word-by-word morpheme annotation, concluded by an English translation. The number of source-gloss line pairs per example varies with sentence length, and these components are not always visually delimited. We extract these triplets using a two-step process:

\begin{itemize}
    \item First convert PDF text into individual text lines ordered top-to-bottom by their vertical coordinates. A single sentence may span multiple lines; this stage operates on raw lines, not linguistic units.
    \item Then fine-tune a BERT\footnote{\texttt{bert-base-multilingual-cased}} classifier \citep{BERT} on $\sim200$ manually annotated examples per language to identify source, gloss, and translation lines.\footnote{Classifier hyperparameters: learning rate = $2 \times 10^{-5}$, 5 epochs, batch size = 16.} Consecutive lines assigned the same label are then grouped to reconstruct complete source sentences, glosses, and translations. We apply rule-based post-corrections exploiting the expected triplet structure to fix occasional misclassifications.
\end{itemize}

The classifier takes each candidate line with a context window of $\pm$3 surrounding lines. Input features include both textual context and geometric properties (bounding box coordinates, line width, aspect ratio). After extraction, we parse English translations with spaCy (\texttt{en\_core\_web\_sm})~\citep{spacy2020} to obtain morphosyntactic metadata (e.g., tense, number, part-of-speech tags).

\paragraph{Section and Dictionary Extraction.} To isolate grammatical content, we implement a heuristic algorithm that detects section headers based on font size statistics, identifying the top-3 distinct sizes as hierarchical levels. Non-grammatical sections (e.g., ``Acknowledgments'', ``History of the community'') are filtered using embedding-based semantic similarity: we compute cosine similarity between section headers and a taxonomy of target categories (e.g., ``Morphology'', ``Syntax'') using \texttt{all-MiniLM-L6-v2} sentence embeddings \citep{sentence-transformers_all-MiniLM-L6-v2_2025}. We retain only sections whose similarity scores exceed the median score across all candidate sections. Additionally, word lists are semi-automatically extracted from glossary sections identified via embedding-based keyword search (``word-list,'' ``dictionary''). We de-duplicate entries and annotate them with part-of-speech tags using spaCy. 

\begin{table}[t]
\centering
\small
\setlength{\tabcolsep}{4pt}
\begin{tabular}{@{}lrrrrrr@{}}
\toprule
& & \textbf{Parallel} & \textbf{Morph} & & \textbf{Sent.} & \textbf{Word} \\
\textbf{Language} & \textbf{Dict} & \textbf{Sent.} & \textbf{Rules} & \textbf{TTR} & \textbf{Len} & \textbf{Len} \\
\midrule
Kalamang & 2,558 & 761 & 399 & 0.36 & 6.3 & 4.5 \\
Tuatschin & 1,788 & 981 & 144 & 0.24 & 11.6 & 3.8 \\
Mandan & 459 & 921 & 519 & 0.64 & 4.0 & 7.0 \\
\bottomrule
\end{tabular}
\caption{Extracted information and corpus statistics based on type-token ratio (TTR), lengths measured in words and characters. Mandan's high TTR (0.64) and word length (7.0) reflect on its polysynthetic morphology.}
\label{tab:corpus_stats}
\end{table}

\paragraph{Rule Extraction and Codification.} To convert narrative grammatical descriptions into codified rules, we use a two-stage extraction process with Gemini-2.5-flash~\citep{gemini_2023}. First, sections exceeding 500 words are semantically chunked by detecting topic shifts via sentence embedding similarity. We set this threshold to balance chunk coherence with sufficient context for rule extraction. Gemini is then prompted to extract rules conforming to the UniMorph annotation schema \citep{batsuren-etal-2022-unimorph}. Each rule is a structured YAML object containing: target part-of-speech, affix type (e.g., suffix, clitic), UniMorph feature-value pairs (e.g., \texttt{CASE:ACC}), and context dependencies (e.g., ``applies to consonant-final stems'').

Second, extracted rules are de-duplicated and converted into deterministic pseudo-code functions (\texttt{ApplyRule(STEM, POS)}) to ensure unambiguous application during synthesis. This codification step consolidates multiple descriptions of the same rule into a single canonical representation. The results of this stage are summarized in Table~\ref{tab:corpus_stats}.

\subsection{Synthetic Corpus Construction}

Relying on the extracted information, we synthetically generate sentences via constrained lexical substitution considering three factors illustrated in Figure~\ref{fig:factorial-design}:

\begin{figure}[ht]
\centering
\resizebox{1\columnwidth}{!}{%
\begin{tikzpicture}[
    every node/.style={font=\small},
    rootnode/.style={rounded corners=3pt, fill=black!80, text=white, font=\small\bfseries, inner sep=5pt},
    posnode/.style={rounded corners=3pt, fill=red!65!black, text=white, font=\footnotesize\bfseries, inner sep=4pt, minimum width=1.1cm},
    grannode/.style={rounded corners=3pt, text=white, font=\scriptsize, inner sep=3pt, minimum width=0.85cm},
    rulestyle/.style={grannode, fill=blue!60!black},
    secstyle/.style={grannode, fill=purple!60!black},
    edgestyle/.style={draw=black!25, thick, -stealth},
    annot/.style={font=\scriptsize\itshape\bfseries, align=center},
]

\foreach \y in {3.5, 2.0, 0.5} {
    \draw[black!8, dashed] (-0.3,\y) -- (8.8,\y);
}

\node[annot, text=red!65!black]  at (-1.3, 3.5) {Target POS\\(4 categories)};
\node[annot, text=blue!60!black] at (-1.3, 2.0) {Retrieval\\Granularity\\(2 levels)};
\node[annot, text=green!50!black] at (-1.3, 0.5) {Sample\\volume, $k$\\(4 levels)};

\node[rootnode] (root) at (4.5, 5.2) {Extracted Parallel Sentences (${\sim}$400)};

\node[font=\scriptsize\itshape, text=black!35, align=center] at (0.0, 4.45) {Filter by\\replaceable\\word of\\target POS};

\node[posnode] (noun) at (1.0, 3.5) {Noun};
\node[posnode] (verb) at (3.5, 3.5) {Verb};
\node[posnode] (adj)  at (5.5, 3.5) {Adjective};
\node[posnode] (adv)  at (8.0, 3.5) {Adverb};

\foreach \pos in {noun, verb, adj, adv}
    \draw[edgestyle] (root.south) -- (\pos.north);

\foreach \pos/\px in {noun/1.0, verb/3.5, adj/5.5, adv/8.0} {
    \node[rulestyle] (\pos-rule) at (\px-0.5, 2.0) {Rule};
    \node[secstyle] (\pos-sec)  at (\px+0.5, 2.0) {Section};
    \draw[edgestyle] (\pos.south) -- (\pos-rule.north);
    \draw[edgestyle] (\pos.south) -- (\pos-sec.north);
}

\foreach \pos/\px in {noun/1.0, verb/3.5, adj/5.5, adv/8.0} {
    \draw[edgestyle] (\pos-rule.south) -- (\px-0.5, 0.72);
    \draw[edgestyle] (\pos-sec.south) -- (\px+0.5, 0.72);
}

\node[rounded corners=3pt, fill=green!50!black,
      text=white, font=\small\bfseries,
      inner sep=5pt, minimum width=8.2cm, minimum height=0.45cm,
      fill opacity=0.75, text opacity=1] at (4.5, 0.5)
      {$k \in \{5, 10, 15, 20\}$ \;\; sentences per combination};

\node[rounded corners=3pt, draw=black!25, fill=black!3, inner sep=4pt, font=\small] at (4.5, -0.4)
    {$4 \times 2 \times 4 = 32$ configurations per language $\times$ 3 languages = 96 experiments};

\end{tikzpicture}
}
\caption{Experimental design. The seed corpus is filtered by target part-of-speech to select sentences containing a replaceable word of that category. Each subset is varied along retrieval granularity and sample volume $k$. All 32 configurations are evaluated independently per language.}
\label{fig:factorial-design}
\end{figure}
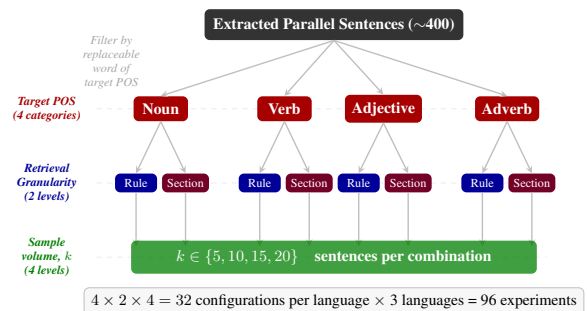

\paragraph{Factor 1: Part-of-Speech (POS).} For each parallel sentence, we identify a replaceable content word belonging to an open POS class, i.e. noun, verb, adjective, or adverb, using greedy string matching against the dictionary, disambiguated via alignment with spaCy-parsed English glosses. Sentences with no dictionary match are excluded from that POS configuration. We then retrieve replacement words from the dictionary subset matching the target POS, ranked by cosine similarity between \texttt{all-MiniLM-L6-v2} embeddings of English definitions. POS tags are assigned by running spaCy on each dictionary entry's English definition in isolation; for single-word definitions this amounts to tagging without 
sentential context, which introduces occasional
mis-tagging. (See Appendix ~\ref{sec:dict_processing} for details).

\paragraph{Factor 2: Retrieval Granularity.} For each substitution, we retrieve relevant grammatical rules using a weighted combination of embedding-based semantic similarity (\texttt{all-MiniLM-L6-v2} \citep{sentence-transformers_all-MiniLM-L6-v2_2025}) and structured feature alignment:
\begin{equation}
S_{\text{final}} = 0.7 \cdot S_{\text{sem}} + 0.3 \cdot S_{\text{struct}}
\end{equation}
where $S_{\text{sem}}$ is the cosine similarity between the query (source sentence plus morphological gloss) and rule embeddings, and $S_{\text{struct}}$ is the proportion of exact POS and UniMorph feature matches between the query and the rule. The weighting (0.7/0.3) prioritizes semantic relevance and is a fixed hyperparameter that we do not optimize. We hold it constant across all languages. We retrieve the top-5 scoring rules and prompt Gemini-2.5-flash to generate the synthetic sentence by applying the retrieved rules to the replacement word within the original sentence context. We test two retrieval granularities: (a)~codified rules with their textual descriptions, and (b)~full source sections from the grammar book. The former provides compact, structured input; the latter preserves phrasal context and parallel examples that may aid generation. To control computational costs, we limit the parallel sentence examples present in the section to a maximum of two per section.

\paragraph{Factor 3: Sample Volume ($k$).} Each seed sentence can produce multiple synthetic variants by substituting different replacement words. We generate 20 replacements per seed sentence, then construct training sets of size $k \in \{5,10,15,20\}$ by taking the subsets with top-$k$ candidate words ranked by embedding-based semantic similarity with the word to be replaced (calculated using cosine similarity of embeddings using \texttt{all-MiniLM-L6-v2} \citep{sentence-transformers_all-MiniLM-L6-v2_2025}) . The smaller training sets are therefore nested subsets of the larger ones. At $k{=}20$ with 400 seed sentences, this targets 8,000 synthetic pairs per POS and granularity combination; at $k{=}5$, approximately 2,000. After filtering malformed outputs (4.2\% average failure rate), training sets range from approximately 1,900 to 7,660 sentences.

The combination of these three factors in a fully-crossed factorial design  yields an extensive number of samples in 32 configurations per language and enables the examination of the impact of each of the manipulated factors.

\subsection{Selected Languages}
We validate the pipeline on three low-resource languages selected to represent diverse typological profiles and distinct language families. \textbf{Kalamang} (Papuan) is an endangered language from the Karas Islands in Indonesia, characterized by agglutinative morphology \citep{Visser2022}. \textbf{Tuatschin} (Romansh) is a dialect of Sursilvan Romansh spoken in Switzerland that exhibits an analytic structure, with source data consisting primarily of narrative text \citep{Maurer-Cecchini2021}. \textbf{Mandan} (Siouan) is a critically endangered polysynthetic language from Fort Berthold, USA \citep{Kasak2024}, featuring complex morphology with a high verb-to-noun rule ratio. Due to the lack of a glossary in the source grammar, the Mandan lexicon was extracted from the external \textit{Comparative Siouan Dictionary} \citep{csd}.
\section{Experiments}
\label{sec:experiments}

\subsection{Experimental Setup}
\label{sec:experimental_setup}

\paragraph{Baselines.} We compare synthetic data augmentation against five baselines. First, we prompt three foundation models (GPT-4.1, GPT-5.2, Gemini-2.5-flash) in a zero-shot setting without language-specific fine-tuning. Second, we use NLLB-200~\citep{nllb-team-2022-language}, a massively multilingual translation model, with typologically or geographically related proxy languages (Indonesian for Kalamang, Romansh for Tuatschin, Dakota for Mandan). Third, we define Seed-Data Fine-Tuning (SDFT) as our primary control by fine-tuning Gemini-2.5-flash exclusively on the 750--900 authentic parallel sentences extracted from each grammar book. This baseline isolates the contribution of synthetic augmentation relative to the limited available gold-standard data. Note that prior work on grammar book translation \citep{tanzer2024benchmarklearningtranslatenew,
aycock2025llmsreallylearntranslate} uses different test sets and evaluation splits, so their reported scores are not directly comparable to ours.

\paragraph{Fine-tuning.} We fine-tune Gemini-2.5-flash on the generated synthetic corpora for 3 epochs with a LoRA adapter rank of 1 to mitigate overfitting.\footnote{We use the Gemini fine-tuning API with default settings except for adapter size = 1.} Test sets comprise 500 held-out original sentences per language, extracted directly from the grammar book; the remaining ${\sim}$800 sentences serve both as the SDFT training set and as the seed pool for synthetic generation. From this pool, we select 400 seed sentences per configuration. We train separate models for each of the 32 configurations per language (Figure~\ref{fig:factorial-design}).

\paragraph{Metrics.} We evaluate translation quality using BLEU~\citep{10.3115/1073083.1073135}, ChrF and ChrF++~\citep{popovic-2015-chrf,popovic-2017-chrf-plus-plus}. We report all three throughout. While BLEU relies on n-grams, ChrF++ provides more stable character-level discrimination at low performance levels.  We therefore rely on ChrF++ as our primary metric.

\subsection{Baselines vs Synthetic Augmentation}

Table~\ref{tab:baselines} presents translation quality across all baselines and the mean performance of our synthetic augmentation across 32 configurations. Zero-shot approaches yield low absolute scores across all languages. For Kalamang, GPT-5.2 achieves the highest zero-shot ChrF++ of $12.03$, followed by GPT-4.1 at $11.34$ and Gemini-2.5-flash at $11.27$. Tuatschin shows substantially higher baseline scores, likely due to related Romansh data in pre-training corpora, with GPT-4.1 reaching $22.64$ ChrF++. Mandan yields the lowest scores, with all zero-shot models below $11.00$ ChrF++ and BLEU under $1.0$, consistent with its high morphological complexity and absence from pre-training data.

The NLLB-200 model underperforms all three LLMs across all languages, even with typologically motivated proxy languages, consistent with the observation that multilingual transfer provides limited benefit in extremely low-resource settings.

In-context learning (ICL) with grammar book content shows a striking language-dependent pattern. For Kalamang, ICL with full sections achieves $27.76$ ChrF++, outperforming all other methods including our best synthetic configuration ($21.48$). For Tuatschin and Mandan, however, ICL underperforms both zero-shot models and SDFT, suggesting that grammar book context at inference time is most effective when the model has no pre-existing knowledge of the language, or needs to be adapted to a specific dialect.

SDFT shows ChrF++ scores of $12.68$ for Kalamang, $26.39$ for Tuatschin, and $13.09$ for Mandan, and serves as the primary control for evaluating synthetic data contributions. Note that SDFT already substantially outperforms all zero-shot and ICL baselines for Tuatschin and Mandan; improvements over SDFT therefore reflect the added value of synthetic augmentation beyond what real parallel data alone provides. Averaged across all configurations, synthetic augmentation outperforms SDFT on all metrics for Kalamang (ChrF++ 14.71 vs.\ 12.68). For Tuatschin and Mandan, the mean falls below SDFT on surface metrics.

\begin{table}[ht!]
\setlength{\tabcolsep}{2pt}
\centering
\resizebox{1\columnwidth}{!}{%
\begin{tblr}{
  row{1} = {font=\bfseries},
  cell{2}{1} = {c=4}{},
  cell{8}{2} = {font=\bfseries},
  cell{8}{3} = {font=\bfseries},
  cell{8}{4} = {font=\bfseries},
  cell{12}{1} = {c=4}{},
  cell{21}{2} = {font=\bfseries},
  cell{21}{3} = {font=\bfseries},
  cell{21}{4} = {font=\bfseries},
  cell{22}{1} = {c=4}{},
  cell{31}{2} = {font=\bfseries},
  cell{31}{3} = {font=\bfseries},
  cell{31}{4} = {font=\bfseries},
  hline{1,32} = {-}{0.08em},
  hline{2-3,12-13,22-23} = {-}{},
  hline{7,17,27} = {-}{dashed},
  hline{9,19,29} = {-}{dashed},
}
Model                                            & BLEU              & ChrF              & ChrF++             \\
\textbf{~ Language: }\textit{Kalamang $\rightarrow$ English} &                   &                   &                    \\
NLLB                                             & 1.12              & 10.69             & 9.23               \\
Gemini (zero-shot)                               & 0.77              & 12.60             & 11.27              \\
GPT-4.1 (zero-shot)                                         & 1.65              & 13.01             & 11.34              \\
GPT-5.2 (zero-shot)                                        & 1.89              & 13.25             & 12.03              \\
ICL Gemini (rule)                                       & 3.77              & 26.91             & 25.32              \\
ICL Gemini (section)                                    & 5.70              & 29.29             & 27.76              \\
SDFT                                             & 1.48              & 14.03             & 12.68              \\
Ours (mean $\pm$ SE)                             & 2.09$\pm$0.25     & 15.65$\pm$0.78    & 14.71$\pm$0.75     \\
Ours (best config)                               & 4.61              & 22.47             & 21.48              \\
\textbf{~ Language: }\textit{Tuatschin $\rightarrow$ English} &                  &                   &                    \\
NLLB                                             & 7.92              & 13.07             & 12.20              \\
Gemini (zero-shot)                               & 8.32              & 23.37             & 22.28              \\
GPT-4.1 (zero-shot)                                          & 12.81             & 23.48             & 22.64              \\
GPT-5.2 (zero-shot)                                          & 10.39             & 20.47             & 19.58              \\
ICL Gemini (rule)                                       & 7.82              & 22.64             & 21.49              \\
ICL Gemini (section)                                    & 9.07              & 24.59             & 23.48              \\
SDFT                                             & 15.31             & 27.29             & 26.39              \\
Ours (mean $\pm$ SE)                             & 14.90$\pm$1.06    & 26.00$\pm$1.09    & 25.44$\pm$1.05     \\
Ours (best config)                               & 22.21             & 32.33             & 31.68              \\
\textbf{~ Language: }\textit{Mandan $\rightarrow$ English} &                     &                   &                    \\
NLLB                                             & 0.63              & 11.37             & 9.79               \\
Gemini (zero-shot)                               & 0.83              & 11.99             & 10.47              \\
GPT-4.1 (zero-shot)                                             & 0.87              & 12.16             & 10.94              \\
GPT-5.2 (zero-shot)                                             & 0.79              & 12.20             & 10.82              \\
ICL Gemini (rule)                                       & 0.17              & 11.71             & 9.75               \\
ICL Gemini (section)                                    & 0.76              & 13.34             & 11.24              \\
SDFT                                             & 0.83              & 15.02             & 13.09              \\
Ours (mean $\pm$ SE)                             & 0.93$\pm$0.13     & 11.55$\pm$0.65    & 10.44$\pm$0.58     \\
Ours (best config)                               & 2.85              & 17.86             & 16.40              \\
\end{tblr}
}
\caption{Baseline and synthetic augmentation results. Methods
above the dashed line operate at inference time; methods below
fine-tune on extracted or synthetic data. ICL prompts Gemini
with extracted rules or full grammar book sections. SDFT
fine-tunes on extracted parallel sentences only. ``Ours (mean)''
reports the mean $\pm$ SE across all 32 configurations; ``Ours
(best)'' reports the single best configuration. \textbf{Bold}
indicates best performance per language.}
\label{tab:baselines}
\end{table}

\subsection{Fine-tuning}

Fine-tuning on synthetically-augmented corpora shows substantial variation across configurations and languages. Averaged across all 32 configurations per language (see Appendix~\ref{sec:appendix} for the full table), Kalamang shows consistent improvement over SDFT, (mean ChrF++ 14.71 $\pm$0.75 SE, 75\% of configurations above baseline). Tuatschin improves at $k \geq 10$  but is pulled down at$k{=}5$ (mean 25.44 $\pm$1.05, 59\% above). Mandan rarely exceeds baseline on surface metrics (mean 10.44 $\pm$0.58, 16\% above).

We now report best-configuration results to characterize the upper bound of pipeline performance. The three languages exhibit distinct patterns. Kalamang benefits most from synthetic augmentation, with the best configuration (section-level retrieval, adjective-numeral replacements, $k{=}20$) with $\Delta=+8.8$. Tuatschin achieves comparable improvements under rule-level retrieval with verb replacements at $k{=}20$ ($\Delta$=$+5.28$), though its mean across configurations falls slightly below SDFT. Mandan proves most resistant to augmentation: 84\% of configurations underperform SDFT on ChrF++. 

Qualitative analysis shows improvements in specific linguistic phenomena. For Kalamang, synthetic training enables correct reordering of adjective-noun constructions from source SOV patterns to English SVO word order. In Mandan, fine-tuned models produce more accurate lexical choices, correctly identifying specific named entities (e.g., ``Cornsilk'') where baseline models produce generic pronouns. For Tuatschin, models successfully learn to interpret complex auxiliary verb constructions (e.g., passive voice with \textit{vegnir}), though occasional errors arise from literal translation of etymologically transparent compound words (e.g., rendering ``June'' as ``weeder'' due to the stem \textit{zarclar} meaning ``to weed'').

\FloatBarrier
\begin{figure*}[]
\centering
\includegraphics[width=\textwidth]{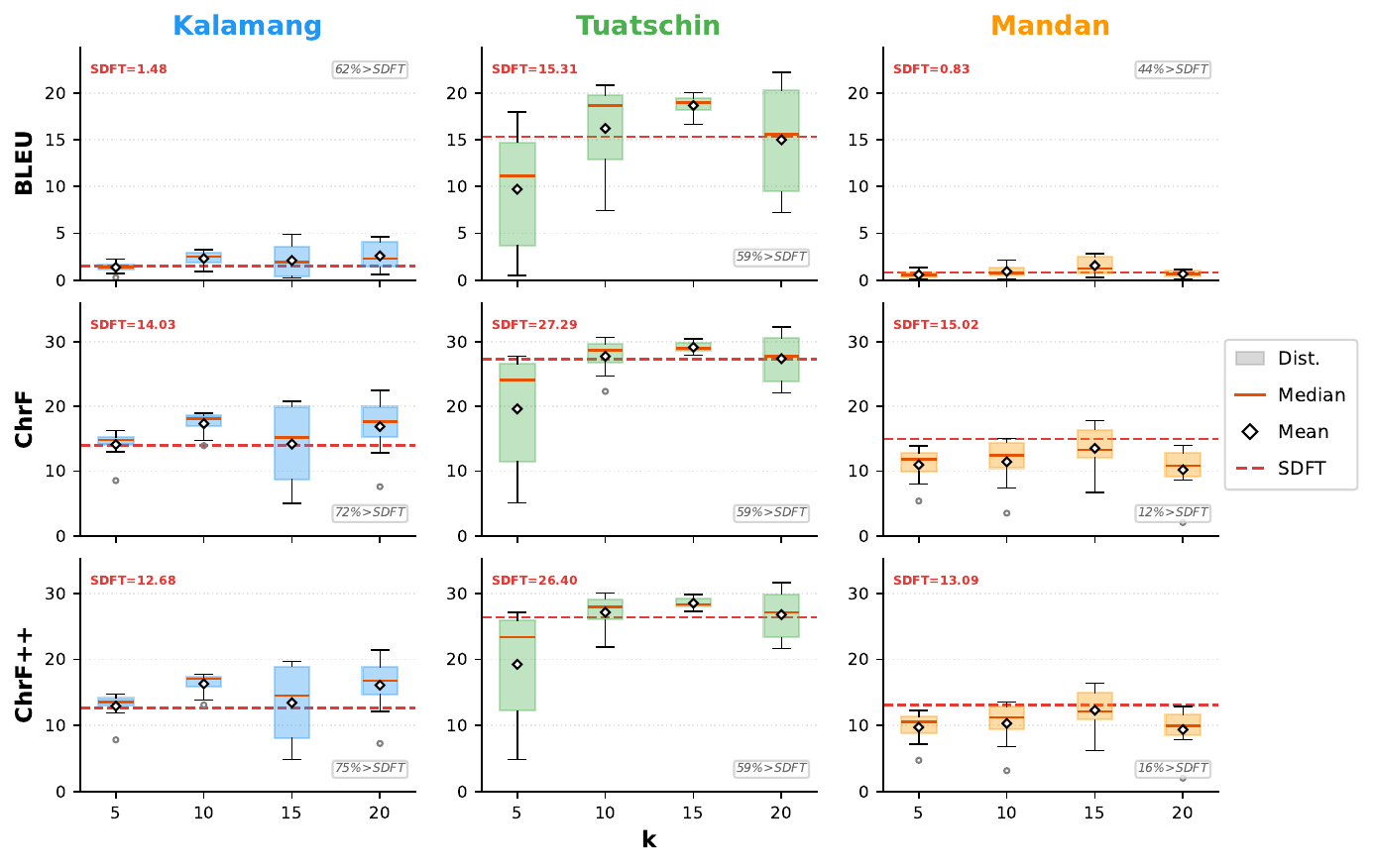}
\caption{Distribution of translation quality across all 32 configurations per language, grouped by sample volume $k$. Red dashed lines mark the SDFT baseline. Kalamang exceeds the baseline at all $k$ values in most configurations. Tuatschin fails at $k{=}5$ but recovers at higher volumes. Mandan rarely exceeds the baseline, with performance peaking at $k{=}15$ before degrading sharply at $k{=}20$.}
\label{fig:results_distribution}
\end{figure*}
\section{Analysis}

Having conditioned the synthetic data generation on the three factors, we can analyze how each of those individually affects translation quality.

\subsection{Impact of Part-of-Speech}

Figure~\ref{fig:pos_comparison} shows the distribution of performance by target POS and retrieval granularity. Averaged across all configurations, Kalamang shows positive mean gains for all POS categories, with nouns producing the most consistent improvement ($\Delta$=+2.98 $\pm$1.06 ChrF++, 75\% above SDFT). For Tuatschin, verbs ($\Delta$=+1.25 $\pm$0.73) and adverbs ($\Delta$=+0.09 $\pm$2.34) are the only categories that reliably improve, while adjective-numeral configurations show high variance ($\Delta$=$-$4.01 $\pm$2.71). Mandan shows negative mean deltas across all POS categories, with nouns performing least poorly ($\Delta$=$-$1.27 $\pm$0.80).

The largest single-configuration improvement is Kalamang's adjective-numeral combination at section-level $k{=}20$ ($\Delta$=+8.80 ChrF++), though the mean for adjective-numeral configurations is more modest ($\Delta$=+2.05 $\pm$1.54). Mandan's adjective-focused configurations are most prone to failure, with only 12\% exceeding the baseline. 

\subsection{Impact of Retrieval Granularity}

\begin{figure*}[h]
\centering
\includegraphics[width=\textwidth]{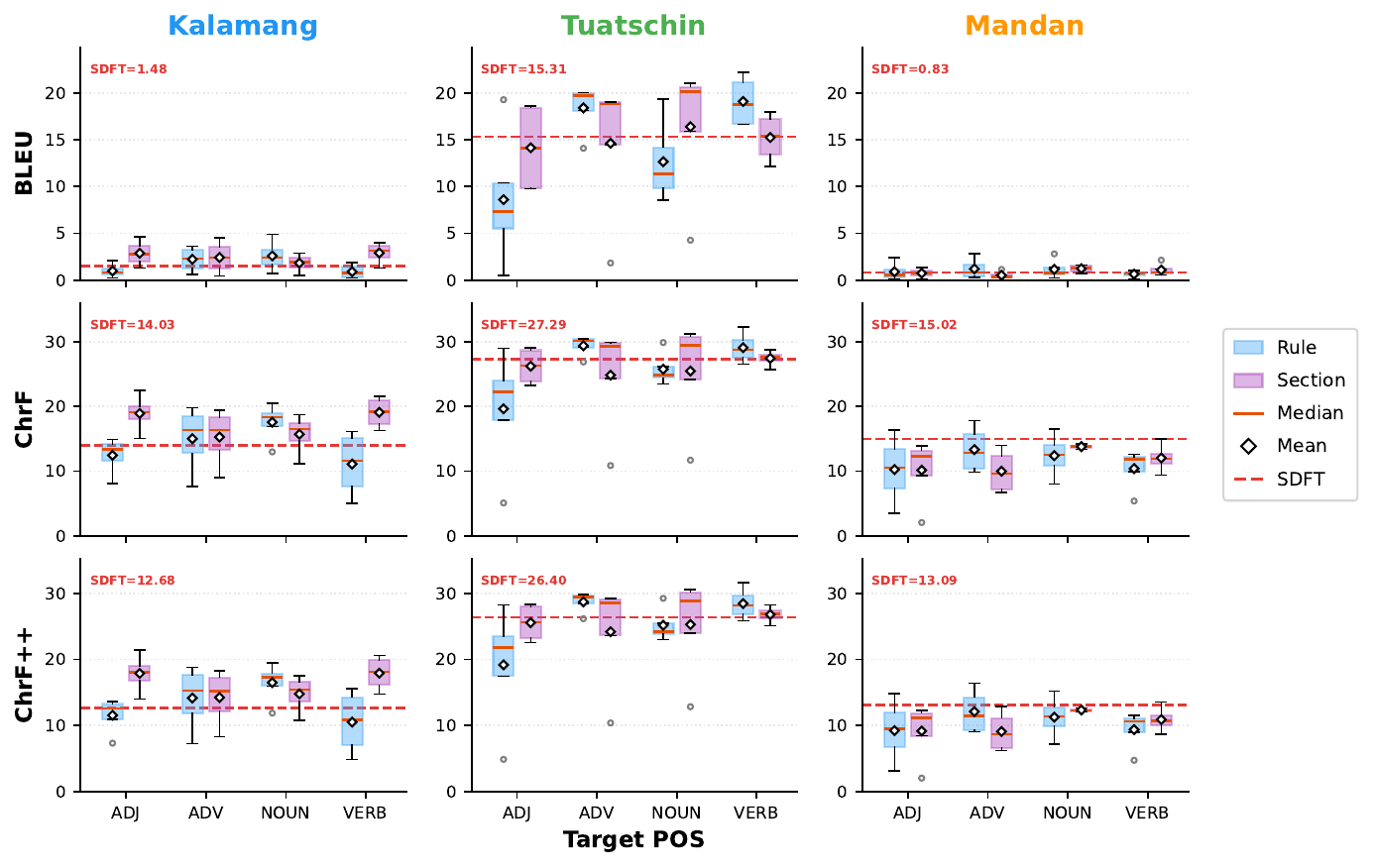}
\caption{Translation quality by target POS and retrieval granularity, aggregated across all sample volumes. For Kalamang, section-level retrieval produces higher medians across most POS categories. For Tuatschin, adjective configurations show high variance while verbs are most stable. Mandan rarely exceeds the SDFT baseline (red dashed lines) regardless of POS or granularity}
\label{fig:pos_comparison}
\end{figure*}

The relative effectiveness of rule-level versus section-level retrieval varies by grammar book, as shown in Figure~\ref{fig:pos_comparison}. For Kalamang, section-level retrieval produces higher mean gains overall ($\Delta$=+3.55$\pm$0.87, 88\% above SDFT) compared to rule-level ($\Delta$=+0.52 $\pm$1.12, 62\% above), with particularly strong advantages for verbs and adjectives. Rule-level retrieval is more effective only for nouns. For Tuatschin and Mandan, the two granularities perform similarly on average. Since we test one grammar book per language, we cannot separate the effect of the language from the effect of how the grammar is written (e.g., whether some sections provide more explicit rule descriptions than others). Full per-POS breakdowns are reported in Appendix~\ref{sec:appendix}.

\subsection{Impact of Sample Volume}

The effect of increasing $k$ depends on both language and configuration, as shown in Figure~\ref{fig:results_distribution}. For Kalamang, mean performance is above baseline at all sample volumes, with the largest gains at $k{=}10$ ($\Delta$=+3.64) and $k{=}20$ ($\Delta$=+3.45). For Tuatschin, $k{=}5$ produces catastrophic failures ($\Delta$=$-$7.13, only 12\% above SDFT), but performance recovers at $k \geq 10$ and peaks at $k{=}15$ ($\Delta$=+2.15, 100\% above). Since smaller training sets are nested subsets of larger ones, the general upward trend from $k{=}5$ to $k{=}15$ is partly expected.

Mandan exhibits non-monotonic behavior, with $k{=}15$ performing least poorly ($\Delta$=$-$0.76, 38\% above) and $k{=}20$ degrading sharply ($\Delta$=$-$3.71, 0\% above). Inspection of the $k{=}20$ outputs shows that the model begins generating grammatical annotations rather than fluent translations, indicating that excessive synthetic data can degrade fine-tuning performance.

\subsection{Interpretation of Results}

The ablation results reveal that synthetic data effectiveness depends on the interaction between language structure and retrieval granularity. 

No single POS category is universally optimal: nouns produce the most consistent gains for Kalamang, verbs for Tuatschin, and no category reliably improves Mandan. Some experimental configurations outperforming SDFT in Tuatschin and Mandan suggests that more comprehensive rule-based generation strategies, beyond single-word lexical substitution, could yield further gains.

Mandan's sharp degradation at $k{=}20$ illustrates a failure mode: when synthetic volume is excessive, the model learns the surface form of training data (interlinear glosses) rather than its function (translation). Inspection of these outputs shows structurally valid but semantically incoherent sentences produced by incompatible feature combinations (e.g., combining a transitive verb suffix with an intransitive stem), indicating that rule retrieval lacks semantic co-occurrence constraints.

We identify three recurring error types across languages. First, models sometimes produce literal translations of compound words instead of recognizing them as expressions. Second, polysynthetic constructions occasionally fragment into unconnected pieces. Third, because synthetic data draws primarily from high-frequency dictionary entries, rare morphological features remain underrepresented. 
\section{Discussion and Conclusion}

We have presented the first semi-automated pipeline for extracting linguistic information from grammar books and synthesizing parallel corpora via rule-guided lexical substitution. Our pipeline operates end-to-end on unstructured PDFs with minimal human intervention. Validated on three typologically diverse languages, the approach achieves a mean ChrF++ improvement of +2.0 for Kalamang across all 32 configurations (75\% above baseline), with best-case gains of +8.8 (Kalamang), +5.3 (Tuatschin), and +3.3 (Mandan). Performance is highly configuration-dependent: Tuatschin improves reliably at $k \geq 10$ but not at $k{=}5$, while Mandan rarely exceeds baseline. The ablation studies further show that configuration factors interact in ways that affect both the magnitude and reliability of gains. 

From a practical standpoint, the pipeline requires only a grammar book PDF as input and produces a synthetic parallel corpus. This makes it applicable to any language with existing grammatical documentation. With over 2,000 endangered languages possessing grammar books but lacking parallel corpora \citep{hammarstrom_endangered}, automated grammar-to-corpus pipelines offer a practical path toward building MT systems to languages that currently have none. Moreover, because the pipeline produces standalone parallel corpora, the generated sentences are not tied to any specific model  architecture. They can equally be used to train smaller, locally deployable MT or NLP systems that do not depend on commercial LLM APIs. It does not replace community-driven language work, but it can provide baseline translation tools that support documentation, education and revitalization efforts.

\section*{Limitations}
Several limitations constrain the generalizability of these results. Most importantly, we rely entirely on automatic metrics; the absence of native-speaker evaluation means we cannot distinguish genuine translation competence from test-set overfitting. Since we test one grammar book per language, observed configuration preferences may reflect properties of how each grammar is written, such as section structure, example density, and rule explicitness, rather than properties of the language itself. The pipeline has been validated on three languages and its performance on tonal systems, non-concatenative morphology, or non-standard grammar book formats remains untested.

Our lexical substitution mechanism lacks semantic co-occurrence constraints, producing sentences that are syntactically valid but sometimes semantically implausible. POS disambiguation is a further source of noise. At the dictionary level, each entry's English definition is tagged by spaCy in isolation; for single-word definitions this amounts to tagging without context. The pipeline currently depends on English-specific NLP tools at multiple stages. This means it supports source-to-English translation directly, but applying it to non-English target languages (e.g., Kalamang-to-Indonesian) would require replacing these components with target-language equivalents, which may not exist for many low-resource settings.

Future work should address these gaps in three directions. First, semantic plausibility filtering, for instance using cross-lingual entailment models to discard incoherent synthetic sentences before training, could reduce the noise that degrades Mandan at high sample volumes. Second, extending the pipeline to the reverse direction (English to low-resource language), validated with native speakers, is necessary for language revitalization applications. Third, replacing the LLM-based extraction stage with smaller fine-tuned models would reduce costs and make the pipeline economically viable at scale. 

\section*{Ethical Considerations}
All grammar books used in this study are published open-access under Creative Commons licenses through Language Science Press. For Kalamang, \citet{tanzer2024benchmarklearningtranslatenew} report that the author obtained community consent for computational use. We have not independently contacted the speech communities for Tuatschin or Mandan. Our pipeline processes only published linguistic descriptions and does not collect new data from speakers. We do not release translation systems, as output quality is not yet sufficient for deployment without community validation.

\section*{Declaration on Generative AI}
During the preparation of this work, the authors used Gemini and ChatGPT for grammar and spelling check. In addition, the authors reviewed and edited the content as needed and take full responsibility for the publication's content.

\section*{Acknowledgments}
This research relies entirely on the foundational work of field linguists and the communities they document. We are deeply indebted to Eline Visser, Ryan Kasak, and Philippe Maurer-Cecchini, whose descriptive grammars of Kalamang, Mandan, and Tuatschin served as the basis of this study. The grammar books used in this study are published open-access under Creative Commons licenses through Language Science Press. We also extend our respect to the speakers of these languages and hope that this work proves beneficial to them. Sina Ahmadi gratefully thanks the support of the UZH Grant (reference number 269093).

\bibliography{references}
\bibliographystyle{acl_natbib}

\clearpage
\appendix

\section{Linguistic Information Extraction}
\label{sec:dict_processing}

Each dictionary entry may contain multiple English definitions (e.g., \textit{yap}: Kalamang entry glossed as both ``black potato'' and ``to divide''). We assign POS tags by running spaCy on each definition independently. When an entry has multiple definitions with conflicting POS tags, we retain all tags and allow the entry to appear in multiple POS-specific candidate pools. For multi-word definitions, spaCy tags the syntactic head word; for single-word definitions, the tag is assigned without sentential context, which can produce errors for ambiguous words (e.g., ``light'' tagged as adjective rather than noun).

During candidate retrieval, we match against the POS of the target word being replaced. If a dictionary entry appears under multiple POS tags, it is only retrieved when its tag matches the target POS for that configuration.

\textbf{Morphosyntactic Feature Extraction.}
\label{app:spacy-fields} 
For each dictionary entry, spaCy processes the English definition and identifies the syntactic head via dependency parsing. The head's POS tag (\texttt{head\_pos}) determines which substitution pool the entry is assigned to: a word is placed in a POS category if at least 50\% of its senses share that head POS. For parallel sentences, spaCy extracts per-token features and aggregates sentence-level properties used for rule selection (Table~\ref{tab:used-features}). Additionally, \texttt{Case} is extracted from the morphological features of the target word being replaced and used for case-specific rule filtering (e.g., locative, accusative).

\begin{table}[!htbp]
\centering
\small

\begin{tabular}{@{}llp{3.8cm}@{}}
\toprule
\textbf{Stage} & \textbf{Feature} &
  \textbf{Usage} \\
\midrule
Dict.\ filtering
  & \texttt{head\_pos}
  & Assigns entries to POS pools \\
\midrule
Rule selection
  & \texttt{tense}
  & Selects tense-related rules \\
  & \texttt{number}
  & Selects agreement rules \\
  & \texttt{gender}
  & Selects gender-related rules \\
  & \texttt{Case}
  & Selects case-specific rules \\
\bottomrule
\end{tabular}
\caption{Morphosyntactic features actively used in the pipeline. All features are extracted by spaCy from English translations.}

\label{tab:used-features}
\end{table}

\textbf{Dictionary Processing Examples.}
\label{sec:spacy_processing}
Table~\ref{tab:dict_examples} illustrates how spaCy processes dictionary entries. For single-word definitions (e.g., \textit{alar} $\rightarrow$ ``fish''), the token is tagged directly as the head. For multi-word definitions (e.g., \textit{emnem} $\rightarrow$ ``old woman''), spaCy identifies the syntactic head (\textit{woman}, NOUN) via dependency parsing; the head POS determines which substitution pool the entry is assigned to.

\begin{table}[!htbp]
\centering
\small

\resizebox{1\linewidth}{!}{%
\begin{tabular}{@{}llllll@{}}
\toprule
\textbf{Lang.} & \textbf{Entry} &
  \textbf{Definition} & \textbf{Head} &
  \textbf{POS} & \textbf{Morph.} \\
\midrule
\multicolumn{6}{@{}l}{\textit{Single-word
  definitions}} \\[2pt]
KLM & \textit{alar}
  & fish & fish & NOUN & Num=Sg \\
TUA & \textit{aua}
  & water & water & NOUN & Num=Sg \\
\midrule
\multicolumn{6}{@{}l}{\textit{Multi-word
  definitions}} \\[2pt]
KLM & \textit{emnem}
  & old woman & woman & NOUN & Num=Sg \\
TUA & \textit{cavagl}
  & horse & horse & NOUN & Num=Sg \\
\bottomrule
\end{tabular}
}
\caption{spaCy analysis of representative dictionary entries. \textsc{Head} indicates the syntactic head used for POS assignment. Single-word entries are tagged in isolation; multi-word entries are parsed for head identification.}
\label{tab:dict_examples}
\end{table}

\section{Detailed Results by Language}
\label{sec:appendix}

This section provides experimental results for all configurations tested across the three target languages. 

Table~\ref{tab:kalamang_results} shows that Kalamang exhibits a split preference for retrieval granularity, with section-level contexts yielding the best overall performance for adjective-numeral replacements at $k{=}20$ (ChrF++ = 21.48, $\Delta$ = +8.8). Rule-level retrieval performed best for noun replacements at $k{=}15$ (ChrF++ = 19.47, $\Delta$ = +6.79).

\begin{table*}[hbtp!]
\centering
\begin{tblr}{
  width = \linewidth,
  colspec = {l l c c c c},
  row{2} = {c},
  cell{1}{1} = {c=3}{c},
  cell{1}{4} = {c=3}{c},
  cell{3}{1} = {r=8}{c,m},
  cell{3}{2} = {r=4}{l,m},
  cell{7}{2} = {r=4}{l,m},
  cell{11}{1} = {r=8}{c,m},
  cell{11}{2} = {r=4}{l,m},
  cell{15}{2} = {r=4}{l,m},
  cell{19}{1} = {r=8}{c,m},
  cell{19}{2} = {r=4}{l,m},
  cell{23}{2} = {r=4}{l,m},
  cell{27}{1} = {r=8}{c,m},
  cell{27}{2} = {r=4}{l,m},
  cell{31}{2} = {r=4}{l,m},
  vline{2} = {1}{},
  vline{4} = {2,4-6,8-10,12-14,16-18,20-22,24-26,28-30,32-34}{},
  vline{3-4} = {1-Z}{},
  hline{1,35} = {-}{0.08em},
  hline{2-3,11,19,27} = {-}{},
  hline{7,15,23,31} = {2-6}{},
}
Language: \textbf{Kalamang $\rightarrow$ English} & & & \textbf{Metrics (Value ($\Delta$))} & & \\
\textbf{POS} & \textbf{Granularity} & \textbf{Samples (k)} & \textbf{BLEU} & \textbf{ChrF} & \textbf{ChrF++} \\
\shortstack[c]{\textbf{ADJ}\\\textbf{+NUM}} & \textbf{rule} & 5 & 2.075 (+0.595) & 14.943 (+0.918) & 13.666 (+0.991) \\
 & & 10 & 0.915 (-0.565) & 13.956 (-0.069) & 13.125 (+0.45) \\
 & & 15 & 0.213 (-1.267) & 8.098 (-5.927) & 7.325 (-5.35) \\
 & & 20 & 0.712 (-0.768) & 12.818 (-1.207) & 12.106 (-0.569) \\
 & \textbf{section} & 5 & 1.293 (-0.187) & 15.095 (+1.07) & 14.04 (+1.365) \\
 & & 10 & 2.246 (+0.766) & 18.992 (+4.967) & 17.803 (+5.128) \\
 & & 15 & 3.337 (+1.857) & 19.196 (+5.171) & 18.271 (+5.596) \\
 & & 20 & 4.614 (+3.134) & \textbf{22.466 (+8.441)} & \textbf{21.475 (+8.8)} \\
\textbf{ADV} & \textbf{rule} & 5 & 1.487 (+0.007) & 14.523 (+0.498) & 13.338 (+0.663) \\
 & & 10 & 3.128 (+1.648) & 18.178 (+4.153) & 17.286 (+4.611) \\
 & & 15 & 3.641 (+2.161) & 19.777 (+5.752) & 18.786 (+6.111) \\
 & & 20 & 0.623 (-0.857) & 7.601 (-6.424) & 7.282 (-5.393) \\
 & \textbf{section} & 5 & 1.507 (+0.027) & 14.66 (+0.635) & 13.46 (+0.785) \\
 & & 10 & 3.252 (+1.772) & 18.041 (+4.016) & 16.952 (+4.277) \\
 & & 15 & 0.419 (-1.061) & 8.982 (-5.043) & 8.332 (-4.343) \\
 & & 20 & 4.505 (+3.025) & 19.419 (+5.394) & 18.28 (+5.605) \\
\textbf{NOUN} & \textbf{rule} & 5 & 0.694 (-0.786) & 12.984 (-1.041) & 11.892 (-0.783) \\
 & & 10 & 2.032 (+0.552) & 18.546 (+4.521) & 17.343 (+4.668) \\
 & & 15 & \textbf{4.891 (+3.411)} & 20.524 (+6.499) & 19.467 (+6.792) \\
 & & 20 & 2.739 (+1.259) & 18.223 (+4.198) & 17.353 (+4.678) \\
 & \textbf{section} & 5 & 2.267 (+0.787) & 15.939 (+1.914) & 14.538 (+1.863) \\
 & & 10 & 2.897 (+1.417) & 18.729 (+4.704) & 17.55 (+4.875) \\
 & & 15 & 0.484 (-0.996) & 11.177 (-2.848) & 10.816 (-1.859) \\
 & & 20 & 1.635 (+0.155) & 17.052 (+3.027) & 16.276 (+3.601) \\
\textbf{VERB} & \textbf{rule} & 5 & 0.268 (-1.212) & 8.538 (-5.487) & 7.835 (-4.84) \\
 & & 10 & 1.251 (-0.229) & 14.747 (+0.722) & 13.838 (+1.163) \\
 & & 15 & 0.215 (-1.265) & 5.034 (-8.991) & 4.822 (-7.853) \\
 & & 20 & 1.87 (+0.39) & 16.101 (+2.076) & 15.591 (+2.916) \\
 & \textbf{section} & 5 & 1.311 (-0.169) & 16.319 (+2.294) & 14.779 (+2.104) \\
 & & 10 & 2.725 (+1.245) & 17.66 (+3.635) & 16.613 (+3.938) \\
 & & 15 & 3.601 (+2.121) & 20.785 (+6.76) & 19.729 (+7.054) \\
 & & 20 & 3.987 (+2.507) & 21.612 (+7.587) & 20.63 (+7.955) \\
\end{tblr}
\caption{Translation performance on Kalamang test sets across all experimental configurations. Results are organized by target POS, retrieval granularity, and sample volume ($k$). Values in parentheses show absolute improvement ($\Delta$) over SDFT baseline. Bold values indicate best performance per metric.}
\label{tab:kalamang_results}
\end{table*}

Table~\ref{tab:tuatschin_results} shows Tuatschin's preference for rule-level retrieval, particularly for verb replacements. The best configuration achieved ChrF++ = 31.68 ($\Delta$ = +5.28) using rule-level retrieval with verbs at $k{=}20$.

\begin{table*}[hbtp!]
\centering
\begin{tblr}{
  width = \linewidth,
  colspec = {l l c c c c},
  row{2} = {c},
  cell{1}{1} = {c=3}{c},
  cell{1}{4} = {c=3}{c}, 
  cell{3}{1} = {r=8}{c,m},
  cell{3}{2} = {r=4}{l,m},
  cell{7}{2} = {r=4}{l,m},
  cell{11}{1} = {r=8}{c,m},
  cell{11}{2} = {r=4}{l,m},
  cell{15}{2} = {r=4}{l,m},
  cell{19}{1} = {r=8}{c,m},
  cell{19}{2} = {r=4}{l,m},
  cell{23}{2} = {r=4}{l,m},
  cell{27}{1} = {r=8}{c,m},
  cell{27}{2} = {r=4}{l,m},
  cell{31}{2} = {r=4}{l,m},
  vline{2} = {1}{},
  vline{4} = {2,4-6,8-10,12-14,16-18,20-22,24-26,28-30,32-34}{},
  vline{3-4} = {1-Z}{},
  hline{1,35} = {-}{0.08em},
  hline{2-3,11,19,27} = {-}{},
  hline{7,15,23,31} = {2-6}{}, 
}
Language: \textbf{Tuatschin $\rightarrow$ English} & & & \textbf{Metrics (Value ($\Delta$))} & & \\
\textbf{POS} & \textbf{Granularity} & \textbf{Samples (k)} & \textbf{BLEU} & \textbf{ChrF} & \textbf{ChrF++} \\
\shortstack[c]{\textbf{ADJ}\\\textbf{+NUM}} & \textbf{rule} & 5 & 0.505 (-14.804) & 5.08 (-22.211) & 4.875 (-21.522) \\
 & & 10 & 7.418 (-7.891) & 22.349 (-4.942) & 21.909 (-4.488) \\
 & & 15 & 19.29 (+3.981) & 28.993 (+1.702) & 28.305 (+1.908) \\
 & & 20 & 7.2 (-8.109) & 22.139 (-5.152) & 21.705 (-4.692) \\
 & \textbf{section} & 5 & 9.834 (-5.475) & 23.248 (-4.043) & 22.59 (-3.807) \\
 & & 10 & 18.382 (+3.073) & 28.584 (+1.293) & 27.928 (+1.531) \\
 & & 15 & 18.617 (+3.308) & 29.06 (+1.769) & 28.382 (+1.985) \\
 & & 20 & 9.764 (-5.545) & 24.063 (-3.228) & 23.437 (-2.96) \\
\textbf{ADV} & \textbf{rule} & 5 & 14.096 (-1.213) & 26.904 (-0.387) & 26.196 (-0.201) \\
 & & 10 & 19.478 (+4.169) & 29.881 (+2.59) & 29.278 (+2.881) \\
 & & 15 & 20.075 (+4.766) & 30.46 (+3.169) & 29.851 (+3.454) \\
 & & 20 & 20.029 (+4.72) & 30.395 (+3.104) & 29.665 (+3.268) \\
 & \textbf{section} & 5 & 1.831 (-13.478) & 10.87 (-16.421) & 10.392 (-16.005) \\
 & & 10 & 18.994 (+3.685) & 28.783 (+1.492) & 28.073 (+1.676) \\
 & & 15 & 18.675 (+3.366) & 29.96 (+2.669) & 29.27 (+2.873) \\
 & & 20 & 19.023 (+3.714) & 29.819 (+2.528) & 29.142 (+2.745) \\
\textbf{NOUN} & \textbf{rule} & 5 & 12.485 (-2.824) & 24.925 (-2.366) & 24.198 (-2.199) \\
 & & 10 & 10.267 (-5.042) & 24.759 (-2.532) & 24.252 (-2.145) \\
 & & 15 & 19.346 (+4.037) & 29.901 (+2.61) & 29.281 (+2.884) \\
 & & 20 & 8.568 (-6.741) & 23.521 (-3.77) & 23.058 (-3.339) \\
 & \textbf{section} & 5 & 4.27 (-11.039) & 11.705 (-15.586) & 12.855 (-13.542) \\
 & & 10 & 20.545 (+5.236) & 30.709 (+3.418) & 30.072 (+3.675) \\
 & & 15 & 19.763 (+4.454) & 28.314 (+1.023) & 27.714 (+1.317) \\
 & & 20 & 21.048 (+5.739) & 31.226 (+3.935) & 30.605 (+4.208) \\
\textbf{VERB} & \textbf{rule} & 5 & 16.698 (+1.389) & 26.544 (-0.747) & 25.863 (-0.534) \\
 & & 10 & 20.847 (+5.538) & 29.657 (+2.366) & 29.122 (+2.725) \\
 & & 15 & 16.65 (+1.341) & 27.924 (+0.633) & 27.275 (+0.878) \\
 & & 20 & \textbf{22.209 (+6.9)} & \textbf{32.334 (+5.043)} & \textbf{31.677 (+5.28)} \\
 & \textbf{section} & 5 & 18.002 (+2.693) & 27.774 (+0.483) & 27.183 (+0.786) \\
 & & 10 & 13.832 (-1.477) & 27.51 (+0.219) & 26.666 (+0.269) \\
 & & 15 & 16.981 (+1.672) & 28.814 (+1.523) & 28.266 (+1.869) \\
 & & 20 & 12.176 (-3.133) & 25.727 (-1.564) & 25.136 (-1.261) \\
\end{tblr}
\caption{Translation performance on Tuatschin test sets across all experimental configurations. Results are organized by target POS, retrieval granularity, and sample volume ($k$). Values in parentheses show absolute improvement ($\Delta$) over SDFT baseline. Bold values indicate best performance per metric.}
\label{tab:tuatschin_results}
\end{table*}

Table~\ref{tab:mandan_results} illustrates Mandan's high performance variability and strong preference for rule-level retrieval. The best configuration used rule-level retrieval with adverbs at $k{=}15$ (ChrF++ = 16.40, $\Delta$ = +3.31), though performance degraded sharply at $k{=}20$ for most POS categories.
\begin{table*}[hbtp!]
\centering
\begin{tblr}{
  width = \linewidth,
  colspec = {l l c c c c},
  row{2} = {c},
  cell{1}{1} = {c=3}{c},
  cell{1}{4} = {c=3}{c}, 
  cell{3}{1} = {r=8}{c,m},
  cell{3}{2} = {r=4}{l,m},
  cell{7}{2} = {r=4}{l,m},
  cell{11}{1} = {r=8}{c,m},
  cell{11}{2} = {r=4}{l,m},
  cell{15}{2} = {r=4}{l,m},
  cell{19}{1} = {r=8}{c,m},
  cell{19}{2} = {r=4}{l,m},
  cell{23}{2} = {r=4}{l,m},
  cell{27}{1} = {r=8}{c,m},
  cell{27}{2} = {r=4}{l,m},
  cell{31}{2} = {r=4}{l,m},
  vline{2} = {1}{},
  vline{4} = {2,4-6,8-10,12-14,16-18,20-22,24-26,28-30,32-34}{},
  vline{3-4} = {1-Z}{},
  hline{1,35} = {-}{0.08em},
  hline{2-3,11,19,27} = {-}{},
  hline{7,15,23,31} = {2-6}{}, 
}
Language: \textbf{Mandan $\rightarrow$ English} & & & \textbf{Metrics (Value ($\Delta$))} & & \\
\textbf{POS} & \textbf{Granularity} & \textbf{Samples (k)} & \textbf{BLEU} & \textbf{ChrF} & \textbf{ChrF++} \\
\shortstack[c]{\textbf{ADJ}\\\textbf{+NUM}} & \textbf{rule} & 5 & 0.746 (-0.081) & 12.453 (-2.569) & 11.075 (-2.014) \\
 & & 10 & 0.074 (-0.753) & 3.499 (-11.523) & 3.16 (-9.929) \\
 & & 15 & 2.42 (+1.593) & 16.387 (+1.365) & 14.819 (+1.73) \\
 & & 20 & 0.324 (-0.503) & 8.659 (-6.363) & 7.889 (-5.2) \\
 & \textbf{section} & 5 & 1.34 (+0.513) & 13.942 (-1.08) & 12.296 (-0.793) \\
 & & 10 & 0.62 (-0.207) & 11.693 (-3.329) & 10.579 (-2.51) \\
 & & 15 & 0.885 (+0.058) & 12.933 (-2.089) & 11.753 (-1.336) \\
 & & 20 & 0.083 (-0.744) & 2.053 (-12.969) & 2.033 (-11.056) \\
\textbf{ADV} & \textbf{rule} & 5 & 0.307 (-0.52) & 10.593 (-4.429) & 9.387 (-3.702) \\
 & & 10 & 1.223 (+0.396) & 15.031 (+0.009) & 13.576 (+0.487) \\
 & & 15 & \textbf{2.847 (+2.02)} & \textbf{17.856 (+2.834)} & \textbf{16.395 (+3.306)} \\
 & & 20 & 0.425 (-0.402) & 9.853 (-5.169) & 9.094 (-3.995) \\
 & \textbf{section} & 5 & 0.363 (-0.464) & 11.842 (-3.18) & 10.57 (-2.519) \\
 & & 10 & 0.259 (-0.568) & 7.409 (-7.613) & 6.767 (-6.322) \\
 & & 15 & 0.299 (-0.528) & 6.69 (-8.332) & 6.186 (-6.903) \\
 & & 20 & 1.144 (+0.317) & 14.028 (-0.994) & 12.86 (-0.229) \\
\textbf{NOUN} & \textbf{rule} & 5 & 0.22 (-0.607) & 8.032 (-6.99) & 7.152 (-5.937) \\
 & & 10 & 0.835 (+0.008) & 13.281 (-1.741) & 11.958 (-1.131) \\
 & & 15 & 2.806 (+1.979) & 16.529 (+1.507) & 15.237 (+2.148) \\
 & & 20 & 0.727 (-0.1) & 11.776 (-3.246) & 10.764 (-2.325) \\
 & \textbf{section} & 5 & 0.704 (-0.123) & 13.86 (-1.162) & 12.129 (-0.96) \\
 & & 10 & 1.585 (+0.758) & 14.221 (-0.801) & 12.642 (-0.447) \\
 & & 15 & 1.602 (+0.775) & 13.741 (-1.281) & 12.46 (-0.629) \\
 & & 20 & 1.037 (+0.21) & 13.372 (-1.65) & 12.18 (-0.909) \\
\textbf{VERB} & \textbf{rule} & 5 & 0.109 (-0.718) & 5.382 (-9.64) & 4.733 (-8.356) \\
 & & 10 & 0.665 (-0.162) & 11.489 (-3.533) & 10.369 (-2.72) \\
 & & 15 & 0.76 (-0.067) & 12.133 (-2.889) & 10.907 (-2.182) \\
 & & 20 & 1.024 (+0.197) & 12.607 (-2.415) & 11.519 (-1.57) \\
 & \textbf{section} & 5 & 0.893 (+0.066) & 11.815 (-3.207) & 10.515 (-2.574) \\
 & & 10 & 2.132 (+1.305) & 14.979 (-0.043) & 13.557 (+0.468) \\
 & & 15 & 0.759 (-0.068) & 11.996 (-3.026) & 10.846 (-2.243) \\
 & & 20 & 0.534 (-0.293) & 9.361 (-5.661) & 8.692 (-4.397) \\
\end{tblr}
\caption{Translation performance on Mandan test sets across all experimental configurations. Results are organized by target POS, retrieval granularity, and sample volume ($k$). Values in parentheses show absolute improvement ($\Delta$) over SDFT baseline. Bold values indicate best performance per metric.}
\label{tab:mandan_results}
\end{table*}

\clearpage
\section{Qualitative results}

\paragraph{Kalamang.}
\label{app:qual:kalamang}
Table~\ref{tab:qual:kalamang} presents eight translation examples from the best-performing Kalamang model. The outputs show that synthetic training enabled correct adjective-noun reordering from source to English (e.g., example~1). The dominant failure mode is hallucination under lexical uncertainty: when the model encounters unknown vocabulary or missing arguments, it repeats salient words (example~2) or inserts entirely unrelated content (examples~3 and~5).

\paragraph{Tuatschin.}
\label{app:qual:tuatschin}
Table~\ref{tab:qual:tuatschin} presents eight translation examples from the best-performing Tuatschin model. The outputs demonstrate robust handling of complex auxiliary constructions: example~1 correctly interprets the passive formed with \textit{vegnir} (``to come'') rather than producing a literal translation of movement, and example~2 is nearly fluent. A recurring failure mode is the literal translation of etymologically transparent compounds: in example~3, the model renders \textit{zarcladur} (``June'', derived from \textit{zarclar}, ``to weed'') as ``weeder'' instead of recognizing it as a fixed calendar term. Example~5 exhibits task confusion, where the model outputs a grammatical analysis of the verb (``1st Person Singular Past Tense'') rather than a fluent translation.

\paragraph{Mandan.} Table~\ref{tab:qual:mandan} presents three translation examples from the best-performing Mandan model. Mandan proved the most difficult target, and the outputs reflect this: example~3 shows the model hallucinating an entirely unrelated narrative scene rather than translating the input. Despite these  failures, the synthetic training data taught the model specific named entities: in example~1, the model correctly identifies \textit{Paxirúuke} as ``Cornsilk'' where the reference translation uses only a generic pronoun. Example~2 illustrates a failure induced by noise in the parallel sentence data, where interlinear gloss tags (\texttt{1sg.poss-pro}) appear directly in the English output instead of fluent text.

\section{Data Synthesis Examples}
\label{app:synthesis-examples}

This appendix illustrates the synthesis pipeline for each language's best-performing configuration. For each example, we show the original sentence, the replacement target, and a sample of generated outputs. Table~\ref{tab:synth-all} presents representative outputs; Table~\ref{tab:rules-kalamang} shows the candidate rules provided to the LLM for one example.

\begin{table}[!ht]
\centering
\small
\begin{tabular}{@{}lp{5.5cm}@{}}
\toprule
\multicolumn{2}{@{}l}{\textbf{Kalamang → English:} ADJ,
  section-level, $k{=}20$} \\
\multicolumn{2}{@{}l}{\textit{Original:}
  \texttt{mu pas sem=ten=at koup}} \\
\multicolumn{2}{@{}l}{\textit{Translation:}
  ``They hugged the scared woman.''} \\
\midrule
\textbf{tayuon} &
  mu pas sem=\textbf{tayuon}=at koup \\
  & $\rightarrow$ They hugged the not-good woman. \\
\textbf{bes} &
  mu pas sem=\textbf{bes}=at koup \\
  & $\rightarrow$ They hugged the good woman. \\
\textbf{siktaktak} &
  mu pas sem=\textbf{siktaktak}=at koup \\
  & $\rightarrow$ They hugged the slow woman. \\
\midrule
\multicolumn{2}{@{}l}{\textbf{Tuatschin → English:} VERB,
  rule-level, $k{=}20$} \\
\multicolumn{2}{@{}l}{\textit{Original:}
  \texttt{Nossad\`{u}na l\`{e}va d\'{a} li
  gjuven pr\'{e}jr...}} \\
\multicolumn{2}{@{}l}{\textit{Translation:}
  ``The Holy Virgin wanted to give...''} \\
\midrule
\textbf{l\`{e}va} &
  ...l\`{e}va d\'{a} li gjuven pr\'{e}jr... \\
  & $\rightarrow$ ...wanted to give the young
  priest... \\
\textbf{sapr\`{e}ndev} &
  ...sapr\`{e}ndev ajn d\'{a} li gjuven... \\
  & $\rightarrow$ ...took herself to give the
  young priest... \\
\textbf{vulaj\`{e}v} &
  ...vulaj\`{e}v d\'{a} li gjuven pr\'{e}jr... \\
  & $\rightarrow$ ...wanted to give the young
  priest... \\
\midrule
\multicolumn{2}{@{}l}{\textbf{Mandan → English:} ADV,
  rule-level, $k{=}15$} \\
\midrule
& r\k{a}\'{h}\v{s}it pxinaate\textipa{P}\v{s} \\
  & $\rightarrow$ He sneezed in back of. \\
& w\k{i}r\k{i}- w\textglotstop{}ahka \\
  & $\rightarrow$ I (acting on you) sneeze. \\
\bottomrule
\end{tabular}
\caption{Representative synthesis outputs for each language's best-performing configuration. Bold marks the substituted
element.}
\label{tab:synth-all}
\end{table}

\begin{table}[!ht]
\centering
\small

\begin{tabular}{@{}clp{4cm}@{}}
\toprule
\textbf{\#} & \textbf{Affix} &
  \textbf{Description} \\
\midrule
1 & \texttt{=at} & OBJ case clitic on NOUN \\
2 & \texttt{=ko} & LOC case on demonstrative ADJ \\
3 & \texttt{=ka} & LAV case on interrogative ADJ \\
4 & \texttt{=saet} & Superlative degree on ADJ \\
5 & \texttt{=ka} & ALL case on demonstrative ADJ \\
\bottomrule
\end{tabular}
\caption{Candidate rules provided to the LLM for the Kalamang ADJ synthesis example above. Rules are retrieved via the scoring function (Eq.~1) and formatted as structured YAML.} \label{tab:rules-kalamang}
\end{table}

\begin{table*}[]
\centering
\small
\begin{tabular}{p{4.5cm} p{5.0cm} p{5.0cm}}
\toprule
\textbf{Source} & \textbf{Reference} & \textbf{Model Output} \\
\midrule
\texttt{[ror wa] me tabusik}
  & This tree is short.
  & That tree is small. \\
\addlinespace
\texttt{Kapal cicauna kon}
  & A small ship.
  & One is small, one is small. \\
\addlinespace
\texttt{som-kon oe-te kon{\textasciitilde}kon}
  & Gave each person one, or gave each person one fruit.
  & (The child) has one hundred teeth, or (the child) has new one hundred teeth. \\
\addlinespace
\texttt{ma hukat= bon mia}
  & He is coming with the net.
  & He/she hits the bottom net. \\
\addlinespace
\texttt{ma rap= et eba gier-un iriskap}
  & When he laughs, his teeth are white.
  & He/she/it hits the whatsit, so his/her/its forehead is blue. \\
\addlinespace
\texttt{wa= nan im karuok weinun}
  & [In] this [picture there are] three bananas too.
  & This is like a twenty net. \\
\addlinespace
\texttt{an tok tumun-an=at boubou=nin}
  & I haven't yet bathed my child.
  & I haven't softened my child yet. \\
\addlinespace
\texttt{gier-ca me me ten}
  & That tooth of yours is bad.
  & All the lobsters are dead. \\
\bottomrule
\end{tabular}

\caption{Translation examples for Kalamang → English (best model: ADJ-NUM, section-level, $k{=}20$).}
\label{tab:qual:kalamang}
\end{table*}
\begin{table*}[]
\centering
\small

\begin{tabular}{p{4.5cm} p{5.0cm} p{5.0cm}}
\toprule
\textbf{Source} & \textbf{Reference} & \textbf{Model Output} \\
\midrule
\texttt{La mùma è vegnida dada in bi schénghètg}
  & Mother was given a beautiful present.
  & *The mother received a nice little present. \\
\addlinespace
\texttt{Quaj taur è bjals, ábar gròns sè 'l bétg}
  & This bull is beautiful, but big he is not.
  & This bull is beautiful, but he is not big. \\
\addlinespace
\texttt{La scùla finéva \ldots\ al matg \ldots\ zarcladur}
  & School ended in May \ldots\ June.
  & School ended \ldots\ in May \ldots\ weeder. \\
\addlinespace
\texttt{1920 [f4] vas schòn a paj}
  & [that's what it takes you to] go on foot.
  & 1920 [f4] you are already paying. \\
\addlinespace
\texttt{L' autar dé va ju gju la lubiantscha dad}
  & I should go and make an appointment with those up there\ldots
  & The original verb form is ``gju'', identified as 1st Person Singular Past Tense\ldots \\
\addlinespace
\texttt{Ju sùn dada gjù séla fatscha}
  & I fell down on my face.
  & I am given down faces. \\
\addlinespace
\texttt{Las tgautschas fòn bétg agl ùm}
  & Trousers do not make a man.
  & My legs don't make the man. \\
\addlinespace
\texttt{Las nòtízjas sa ju bétg danùndar als gjaniturs}
  & I don't know where my parents had the news from.
  & The news I don't know where my parents had them from. \\
\bottomrule
\end{tabular}
\caption{Translation examples for Tuatschin → English (best model: VERB, rule-level, $k{=}20$).}
\label{tab:qual:tuatschin}

\end{table*}

\begin{table*}[]
\centering
\small

\begin{tabular}{p{4.5cm} p{5.0cm} p{5.0cm}}
\toprule
\textbf{Source} & \textbf{Reference} & \textbf{Model Output} \\
\midrule
\texttt{Paxirúukeena miní hįįhere}
  & He was drinking water.
  & Cornsilk made him drink water. \\
\addlinespace
\texttt{Mí'shak máa'ąk íwasek pshíiwahaani\ldots}
  & My land\ldots
  & \texttt{1sg.poss-pro} land\ldots \\
\addlinespace
\texttt{Kináatani máapte kaní'ni kasími réehoomako'sh}
  & All around, there were very thick tall trees.
  & He got up again, climbed up the river bank, and he set off traveling. \\
\bottomrule
\end{tabular}
\caption{Translation examples for Mandan → English (best model: ADJ-NUM, rule-level, $k{=}15$).}
\label{tab:qual:mandan}
\end{table*}

\clearpage
\section{Prompts}

This section contains all the prompt templates used. The rule extraction prompt is given in Figure \ref{fig:rule-extraction-prompt}. The output of these rules were cleaned, deduplicated and codified using the prompt in Figure \ref{fig:codification-prompt}.

\begin{figure}[!htbp]
\small
\begin{tcolorbox}[colback=gray!5, colframe=gray!60,
  boxrule=0.5pt, arc=2pt, left=4pt, right=4pt,
  top=2pt, bottom=2pt]
\ttfamily
You are an expert descriptive linguist and data
structurer. Your core task is to \textbf{extract
morphological and syntactic rules} from a grammar
book paragraph for a low-resource language.
These rules will be used by a downstream AI agent
to synthesize valid sentences.\\[4pt]
\textbf{Selection Criteria}\\
Extract a rule IF AND ONLY IF:\\
1. It involves an \textbf{overt surface change}
(suffix, prefix, clitic, particle, or mutation).\\
2. It is \textbf{productive} (applies generally to
a class of words, not one specific exception).\\
3. It has a clear \textbf{grammatical function}
(Case, Tense, Mood, Aspect, Agreement).\\
Do NOT extract: purely phonological rules,
typological trivia, examples, or footnotes.\\[4pt]
\textbf{Output Schema (YAML)}\\
\texttt{category: surface\_rule}\\
\texttt{description:} \textit{<POS, morpheme, and grammatical function>}\\
\texttt{target\_pos:} \textit{<NOUN | VERB | ADJ | PRON>}\\
\texttt{affix\_type:} \textit{<SUFFIX | PREFIX | CLITIC | PARTICLE | ...>}\\
\texttt{morpheme:} \textit{<literal string, e.g., "-at", "ko=">}\\
\texttt{application\_string:} \textit{<e.g., STEM + "-an">}\\
\texttt{unimorph\_feature:} \textit{<e.g., CASE, TENSE, AGR, NUM>}\\
\texttt{unimorph\_value:} \textit{<e.g., ACC, PST, PL, 3SG>}\\
\texttt{context\_dependency:} \textit{<phonological/agreement environment, or N/A>}\\
\texttt{semantic\_trigger:} \textit{<when to apply this rule>}\\[4pt]
\textbf{Output Handling}\\
If no rule is found: output \texttt{category: N/A} and
\texttt{description: "No applicable rule found."}\\[4pt]
\textbf{Task}\\
Process the following paragraph: \{input\_paragraph\}
\end{tcolorbox}
\caption{Prompt template for grammar rule extraction.
The model outputs one YAML block per extracted surface
rule, or a fixed \texttt{N/A} block if no applicable
rule is found. The \texttt{\{input\_paragraph\}}
placeholder is filled with a single paragraph from
the parsed grammar book.}
\label{fig:rule-extraction-prompt}
\end{figure}

\begin{figure}[!htbp]
\small
\begin{tcolorbox}[colback=gray!5, colframe=gray!60,
  boxrule=0.5pt, arc=2pt, left=4pt, right=4pt,
  top=2pt, bottom=2pt]
\ttfamily
You are a computational linguist.\\
Task: Translate the single input rule into a concise
pseudo-code function, ensuring the output YAML is
perfectly parsable.\\[4pt]
\textbf{Output Schema (YAML)}\\
\texttt{canonical\_description:} \textit{<single sentence synthesizing}\\
\phantom{\texttt{canonical\_description: }}\textit{meaning, usage, and all conditions>}\\
\texttt{lrl\_code:} \textit{<pseudo-code function applying the rule>}\\[4pt]
\textbf{Codification Rules}\\
1. Function MUST be named \texttt{ApplyRule(STEM, POS)}.\\
2. Affixation must use the \texttt{application\_string} field.\\
3. Implement all conditions with \texttt{IF/ELSE IF} checks\\
\phantom{3. }(e.g., \texttt{IF STEM\_ENDS\_IN('x')}, \texttt{IF POS == 'Y'}).\\
4. If conditions are not met, MUST return the original\\
\phantom{4. }\texttt{STEM}.\\[4pt]
\textbf{Input Rule (YAML)}\\
Process this rule.\\
---\\
\{input\_rule\}\\
---
\end{tcolorbox}
\caption{Prompt template for rule codification.
Each extracted surface rule (in YAML) is converted
into a \texttt{canonical\_description} and a
deterministic \texttt{ApplyRule(STEM, POS)}
pseudo-code function. The codified rules are indexed
for semantic retrieval and serve as the grammar
representation in the \textit{rule}-granularity
generation condition.}
\label{fig:codification-prompt}
\end{figure}

\clearpage

The following template (Figure \ref{fig:base-prompt}) is used across all model configurations. For \textit{rule} granularity, \texttt{\{rules\_data\}} contains individual morphological rules (e.g., clitic attachment patterns). For \textit{section} granularity, it contains full grammar book sections.

\begin{figure}[!htbp]
\small
\begin{tcolorbox}[colback=gray!5, colframe=gray!60,
  boxrule=0.5pt, arc=2pt, left=4pt, right=4pt,
  top=2pt, bottom=2pt]
\ttfamily
You are a professional linguist for \{lang\}.\\
Task: Replace the \{pos\} in the sentence using
provided rules. Output must be concise.\\[4pt]
\textbf{Constraints}\\
-- Deduce existing clitics by comparing the
original sentence to the provided stem.\\
-- Allomorphy: Apply the correct variant based on
the new stem's ending.\\
-- You MUST cite Rule \#s for all changes.\\[4pt]
\textbf{Inputs}\\
-- \{granularity\_statement\}:
\{rules\_data\}\\
-- Sentence: \{sentence\_text\}\\
-- Translation: \{translation\}\\
-- Target: Replace \{stem\_to\_replace\}
(\{original\_gloss\_features\}) with \{word\}
("\{word\_translation\}") [POS: \{pos\}]\\[4pt]
\textbf{Step-by-step reasoning}\\
1. Context: [Target word] + [Deduced clitics]\\
2. Rule Selection: [Rule \#s] + [Justification]\\
3. Transformation: [New stem] + [Applied changes]\\
4. Final Sentence: [\{lang\} result]\\
5. Translation: [English result]\\[4pt]
\textbf{Output format}\\
\texttt{final\_sentence: "..."}\\
\texttt{english\_translation: "..."}
\end{tcolorbox}
\caption{Prompt template for grammar-aware
synthetic generation. Placeholders (in braces)
are filled per language and configuration. The
\texttt{granularity\_statement} field contains
either codified rules or full grammar book
sections depending on the retrieval granularity
condition.}
\label{fig:base-prompt}
\end{figure}

The prompt for the ICL baseline is found in Figure \ref{fig:icl-prompt}. The prompt in Figure \ref{fig:finetune-prompt} was used for finetuning, zero-shot inference and final inference from the finetuned (including SDFT) models. 

\begin{figure}[!htbp]
\small
\begin{tcolorbox}[colback=gray!5, colframe=gray!60,
  boxrule=0.5pt, arc=2pt, left=4pt, right=4pt,
  top=2pt, bottom=2pt]
\ttfamily
You are a professional linguist for \{lang\}.\\
Task: Translate the sentence into English using the
provided grammatical rules.
Output must be concise. Use fragments, not full
paragraphs.\\[4pt]
\textbf{Constraints}\\
-- Use the provided rules to interpret morphological
forms in the sentence.\\
-- Use the dictionary entries to identify known word
meanings.\\
-- Citation: You MUST cite Rule \#s used in your
reasoning.\\[4pt]
\textbf{Inputs}\\
-- \{granularity\_statement\}:\\
\{rules\_data\}\\
-- Sentence: \{sentence\_text\}\\
-- Dictionary (source words found in lexicon):\\
\{dictionary\_entries\}\\[4pt]
\textbf{Step-by-step reasoning (Direct \& Brief)}\\
1. Segmentation: [Words + clitics/morphology
identified].\\
2. Rule Selection: [Rule \#s] used + [1-phrase
justification].\\
3. Word Meanings: [Known words from dictionary].\\
4. Translation: [English result].\\[4pt]
\textbf{Output format}\\
\texttt{english\_translation: "..."}
\end{tcolorbox}
\caption{Prompt template for the grammar-aware ICL translation baseline. Unlike the generation prompt (Figure~\ref{fig:base-prompt}), there is no target word to substitute; instead the model translates the full source sentence directly. The \texttt{granularity\_statement} field selects between codified rules (\textit{rule} condition) and raw grammar book paragraphs (\textit{section} condition). Dictionary entries are retrieved by tokenising the source sentence and looking up sub-morphemes in the bilingual lexicon.}
\label{fig:icl-prompt}
\end{figure}

\begin{figure}[!htbp]
\small
\begin{tcolorbox}[colback=gray!5, colframe=gray!60,
  boxrule=0.5pt, arc=2pt, left=4pt, right=4pt,
  top=2pt, bottom=2pt]
\ttfamily
\textbf{User turn}\\
Translate to english from \{language\_name\}:
\{source\_sentence\}\\[6pt]
\textbf{Model turn} (training target only)\\
\{english\_translation\}
\end{tcolorbox}
\caption{Prompt template used for Gemini fine-tuning and baseline batch inference. Each example is formatted as a two-turn conversation: the user provides the source sentence prefixed with a language identifier, and the model response is the English translation. At inference time only the user turn is sent; the model turn is omitted.}
\label{fig:finetune-prompt}
\end{figure}

\end{document}